\documentclass{article}

\usepackage[preprint]{corl_2026} 
\usepackage[ruled,vlined,linesnumbered]{algorithm2e}
\usepackage{booktabs} 
\usepackage{xcolor}
\usepackage{amssymb}
\usepackage{graphicx}
\usepackage{subcaption}
\usepackage{amsmath}
\usepackage{xcolor}

\usepackage{float} 

\usepackage{listings}
\usepackage{adjustbox}   
\lstdefinestyle{rlang}{
  basicstyle=\ttfamily\scriptsize,
  commentstyle=\color{gray}\itshape,
  keywordstyle=\color{blue!70!black}\bfseries,
  morekeywords={def,return,if,else,Feature,Action},
  morecomment=[l]{\#},
  columns=fullflexible,
  keepspaces=true,
  frame=single,
  framerule=0.4pt,
  rulecolor=\color{gray!50},
  backgroundcolor=\color{yellow!8},
  xleftmargin=2pt, xrightmargin=2pt,
  aboveskip=0pt, belowskip=0pt
}

\title{Reinforcement Learning from LLM-Generated (Suboptimal) Demonstrations}

\author{
  Anurag Kadkol \\
  Khoury College of Computer Sciences\\
  Northeastern University\\
  United States\\
  \texttt{kadkol.a@northeastern.edu} \\
  \And
  Lawson L.S. Wong \\
  Khoury College of Computer Sciences\\
  Northeastern University\\
  United States\\
  \texttt{lsw@ccs.neu.edu} \\
}

\begin{document}
\maketitle


\begin{abstract}
    Reinforcement learning (RL) has achieved substantial progress across many domains, but training an agent \textit{tabula rasa} remains highly sample-inefficient. 
    A common remedy is to initialize the agent using a neural network pretrained on demonstrations. However, collecting demonstrations is expensive, and the demonstrations themselves can be suboptimal.
    In this work, we use a large language model (LLM) to generate an advice policy that produces demonstration data with effectively no human cost. We use this data to warmstart a DQN agent, and use the advice policy itself to guide exploration during training. 
    Because LLM-generated advice is often suboptimal, we further introduce a value warmstart method that initializes the agent to recognize promising states from the advice policy's behavior, without directly copying its actions.
    To characterize the robustness of this method, we systematically perturb individual sub-skills of an LLM-generated advice policy for a long-horizon, sparse-reward task. Degraded advice still meaningfully accelerates learning relative to unaided baselines.
    These results suggest that LLM-generated advice, though suboptimal, can be a cheap alternative to costly human demonstrations. 
\end{abstract}



\section{Introduction}

Reinforcement Learning (RL) methods typically learn \textit{tabula rasa} through unaided trial and error. Although they have achieved success in various domains\citep{mnih2015human, berner2019dota, nair2018overcoming} they remain highly sample inefficient, making RL impractical when environment interactions are slow or expensive. A common remedy to learning from scratch is training on examples of desired behavior, allowing agents to learn from existing behavior rather than discovering it. A policy can be trained to directly reproduce demonstrated behavior, or an RL agent can use demonstration data to initialize its learned components, reducing the interactions needed to reach strong performance.

Regardless, these approaches depend on having demonstrations to learn to begin with. Demonstrations, however, are expensive to obtain, typically requiring a human expert to operate the system or label preferred behavior at scale. They are also often imperfect as execution failures, near-misses, and inconsistent behavior are common even in expert-collected data. Suboptimality is therefore a standing property of demonstration data, not a shortcoming specific to any one source, and any method that learns from demonstrations must eventually contend with it.

We address both problems together. We use an LLM to produce an executable \textbf{advice policy}, expressed in RLang\citep{rodriguez2023rlang}, an algorithm- and environment-agnostic DSL for describing structured task knowledge to RL agents. The advice policy provides action recommendations using the agent observations or environment state that serve both as demonstrations for initializing the RL agent and as an exploration aid during training. Rolling out the advice policy produces demonstrations with effectively no human cost. We use this data to warmstart a DQN agent, and continue using the advice policy to guide exploration during training.

However, the advice policy is sometimes imperfect, and directly learning from its actions can also transfer mistakes to the agent. To mitigate this, we further introduce value warmstart, which learns from the advice policy's own induced behavior without directly copying its actions. Specifically, we behavior clone the advice policy, estimate the returns induced by this policy and use the resulting critic to initialize the DQN's Q-network. This allows the agent to learn which states are promising according to the behavior induced by the advice while remaining free to select actions different from those taken by the advice policy.

Across a suite of MiniGrid environments and controlled advice degradation, these methods continue to accelerate learning over an unaided baseline even when the underlying advice is substantially suboptimal.


\section{Related Work}
\label{sec:related}

\textbf{LLMs for Decision Making.} LLMs have been used to generate a range of artifacts for sequential decision-making, including high-level plans \citep{ahn2022saycan}, executable skills \citep{wang2023voyager}, reward functions \citep{ma2023eureka}, exploration goals \citep{du2023guiding}, and auxiliary rewards derived from external knowledge \citep{wu2023read}. In robotics, LLMs have been used to design executable controllers\citep{liang2023code}, task programs\citep{singh2023progprompt} and manipulation maps\citep{huang2023voxposer}. Closest to our setting are \citet{sadmine2024language} and \citet{liu2025synthesizing}, where LLMs generate programmatic policies that seed subsequent search or refinement, similar to our offline advice generation using LLMs.

\textbf{Learning from (Suboptimal) Demonstrations.} Demonstrations have been incorporated in sequential decision making problems. Imitation learning based approaches \citep{pomerleau1991efficient,ross2011reduction} directly try to mimic the demonstrated behavior. Offline RL methods \citep{kumar2020conservative} learn entirely from the fixed demonstration dataset without further environment interaction. Inverse RL algorithms \citep{abbeel2004apprenticeship,ziebart2008maximum} infer the reward from demonstrations. Algorithms like DQfD~\cite{hester2018deep}, DAPG~\cite{rajeswaran2018dapg}, and AWAC~\cite{nair2020awac} use demonstrations to pretrain their neural networks before interacting with the environment. These approaches generally rely on expert demonstrations, which can be expensive to collect; in contrast, we use LLM-generated advice policies as a low-cost source of demonstrations for warmstarting agents.

A related literature considers suboptimal or incomplete demonstrations, including ranked demonstrations \citep{brown2019extrapolating}, quality-weighted imitation \citep{wu2019imitation}, and partial or subgoal-level supervision combined with RL \citep{le2018hierarchical,lynch2019play}. These settings primarily study a dense reward setting where even imperfect demonstrations carry some reward signal. Our setting considers sparse reward tasks where reward is only given on successful termination. Also, we study controlled degradation of advice at the individual sub-skill level, studying the effects of various qualities of degradation.


\section{Background}
\label{sec:background}

\subsection{Preliminaries}
\textbf{POMDP}. The tasks here are modelled as partially observable Markov decision processes (POMDPs), defined by $(S, A, T, R, \Omega, O, \gamma)$, where $S$ is the state space, $A$ is the action space, $T(s'|s, a)$ is the transition function, $\Omega$ is the observation space, $O(o|s', a)$ is the observation function giving the probability of observing o after transitioning to s' and $\gamma$ is the discount factor. The agent doesn't observe the state $s$ directly, only an observation $o \in \Omega$ and must act on the basis of its observation alone. This distinction matters for our method: our advice policies are defined either over observations or the underlying state.

\textbf{DQN}. Deep Q-Network (DQN) approximates the action-value function $Q(s,a;\theta)$ with a neural network, trained to minimize the temporal-difference (TD) loss
$L_{TD} = (y - Q(s,a;\theta))^2$ where $y = r + \gamma max_{a'}Q(s',a', \theta^-)$

using transitions sampled from a replay buffer and a target network $\theta^-$ updated periodically toward $\theta$. We build directly on this objective: our proposed methods pretrain $Q(s,a;\theta)$ before online training begins. 

\section{Method}
\label{sec:method}

\subsection{Generating Advice Policy and Demonstrations}
\label{sec:advice_data_gen}

\textbf{RLang.} Advice policies are expressed in RLang\citep{rodriguez2023rlang}, a declarative, algorithm- and environment-agnostic language for describing partial world knowledge to RL agents. We use two of the available declarations: A \textit{Feature} which computes a Boolean or numeric condition over an observation or state (e.g., whether an object is visible); a \textit{Policy} which selects an action via conditional logic over these features. RLang programs are parsed directly into executable objects with no additional learning required. An example of an LLM-generated RLang program features is shown in Appendix ~\ref{sec:advice_generation_pipeline}.

\textbf{Advice Policy Generation.} To generate an advice policy for a task, we prompt an LLM with RLang's documentation (its grammar and annotated example programs) and a description of the task which includes the task objective, observation/action space and optionally a textual strategy to solve the task. For state-based advice, we additionally provide predicate functions such as $at(location)$ and $go\_to(object)$ as translation targets. The LLM is asked to produce both the Feature declarations needed to ground the task and a Policy declaration that solves it. If the LLM's first attempt fails or produces degenerate behavior, we re-prompt with a natural-language hint; the number of re-prompts varies by task. Producing $\pi_{advice}$ this way costs a small number of LLM queries and no environment interaction. The advice generation pipeline is described in Appendix ~\ref{sec:advice_generation_pipeline}

\textbf{Generating Demonstrations.} The advice policy is rolled out in the environment to collect $(s/o, a, r, s'/o')$ trajectories. This step is cheap relative to having humans collect demonstrations by either manually controlling the agent or handcrafting a policy. Because $\pi_{advice}$ is generated using natural language description of the task and the LLM's own parametric knowledge of the domain, the advice can be suboptimal. The resulting demonstrations can thus exhibit degenerate behavior, such as agents moving in circles or repeating and alternating actions, missing sub-skills, or sequencing sub-skills incorrectly. Examples of such behavior is shown in Appendix ~\ref{sec:obs-based_suboptimal}.

\subsection{Method 1: Warmstarting from Advice Demonstrations}
\label{sec:warmstart_optimal}

\textbf{Warmstarting.}\label{sec:warmstart} We use the generated demonstrations to warmstart a DQN agent before online training begins. Episodes are retained in the replay buffer only if they terminate with a non-zero return. Failed episodes are discarded to prevent mediocre advice policy from poisoning the replay buffer with failed rollouts. Before any online environment interactions, the Q-network is trained offline using a combined 1-step and n-step TD loss:
$L = L_{TD} + \lambda L_{n\_step}$
 where $L_{TD} = \left(y - Q(o,a;\theta)\right)^2$ with $y = r + \gamma\max_{a'} Q(o',a';\theta^-)$, 
 and $L_{n\_step} = \left(y_n - Q(o,a;\theta)\right)^2$ with $y_n = \sum_{k=0}^{n-1}\gamma^k r_{t+k} + \gamma^n\max_{a'} Q(o_{t+n},a';\theta^-)$.

The n-step term provides longer-horizon credit assignment, which is particularly useful under the sparse rewards common to long-horizon object interaction tasks, where the 1-step TD signal is zero for most transitions. This phase gives the Q-network a head start over a randomly initialized network, without requiring any interaction beyond what was already spent generating demonstrations.

\textbf{Exploration Advice}\label{sec:exploration_advice}. Once online training begins, we do not discard the advice policy: standard $\epsilon$-greedy is replaced with a three-way split, where the agent acts randomly with probability $\epsilon/2$, follows $\pi_{advice}$ with probability $\epsilon/2$, and acts greedily otherwise. As $\epsilon$ anneals over training, the advice's influence fades naturally, and the agent increasingly relies on its own learned policy rather than remaining bound to the advice's limitations.

\begin{algorithm}[H]
{\small
    \caption{Warmstarting the Q-Network}
    \label{alg:warmstart}
    \KwIn{Advice policy $\pi_{\text{advice}}$, precollect steps $M$, warmstart steps $G$, discount $\gamma$, $n$-step weight $\lambda$}
    \KwOut{warmstarted Q-network $Q(o,a;\theta)$}
    \BlankLine
    Initialize replay buffer $\mathcal{D}$, Q-network $\theta$, target network $\theta^- \gets \theta$\;
    \BlankLine
    \tcp{Phase 1 -- Precollect: roll out advice, keep only successful episodes}
    \While{$|\mathcal{D}| < M$}{
        Roll out $\pi_{\text{advice}}$ for one episode, storing $(o_t, a_t, r_t, s_{t+1}, d_t)$\;
        \lIf{episode return $> 0$}{add all transitions to $\mathcal{D}$}
        \lElse{discard episode}
    }
    \BlankLine
    \tcp{Phase 2 -- Pretrain on precollected data only, no environment interaction}
    \For{$g = 1$ \KwTo $G$}{
        Sample mini-batch from $\mathcal{D}$\;
        $y \gets r + \gamma(1-d)\max_{a'}\hat{Q}(o',a';\theta^-)$; \quad compute $n$-step target $y_n$\;
        Update $\theta$ by minimizing $L_{\text{TD}} + \lambda L_{n\text{-step}}$\;
        Periodically sync $\theta^- \gets \theta$\;
    }
    \Return{$Q(\cdot,\cdot;\theta)$}
}
\end{algorithm}

\begin{algorithm}[H]
{\small
    \caption{DQN with Exploration Advice}
    \label{alg:exploration_advice}
    \KwIn{Warmstarted $Q(o,a;\theta)$, advice policy $\pi_{\text{advice}}$, exploration schedule $\varepsilon_t$, total steps $T$, exploration fraction $f$}
    \KwOut{Trained Q-network $Q(o,a;\theta)$}
    \BlankLine
    \For{$t = 1$ \KwTo $T$}{
        $\epsilon \gets \max\!\bigl(\epsilon_{\text{end}},\; \epsilon_{\text{start}} - (\epsilon_{\text{start}} - \epsilon_{\text{end}})\,\tfrac{\mathrm{t}}{f \cdot T}\bigr)$\;
 
        \BlankLine
 
        Select $a \gets
        \begin{cases}
            \text{random action}    & {\text{with probability } \epsilon/2} \\
            {\pi_{\text{advice}}(o)} & {\text{with probability } \epsilon/2} \\
            \arg\max_{a}\, Q(o, a;\,\theta)          & \text{otherwise}
        \end{cases}$\;
        Execute $a$, observe $r_t, s', d$; store $(o, a, r_t, o', d)$ in $\mathcal{D}$\;
        Sample mini-batch from $\mathcal{D}$; update $\theta$ via TD loss\;
        Periodically sync $\theta^- \gets \theta$\;
    }
    \Return{$Q(\cdot,\cdot;\theta)$}
}
\end{algorithm}

\subsection{Method 2: Value Warmstarting from Suboptimal Demonstrations}
\label{sec:warmstart_suboptimal}

Warm-starting from advice transitions (Section~\ref{sec:warmstart_optimal}) trains the Q-network only on the actions the advice took. The pretrained greedy policy therefore tends to reproduce those actions, including incorrect advice actions, and must unlearn them online (Appendix~\ref{sec:warmstart_state-based_suboptimal}). TD pretraining also extracts little signal under sparse rewards as mini-batches are sampled as individual transitions. Because only the final transition of a successful episode carries reward, the return must propagate backward a few steps per update. Degraded advice worsens this, since its longer episodes further dilute the rewarded transitions. 
Value Warmstart addresses both problems. Rather than learning which actions the advice takes, it learns which states are promising under the behavior the advice induces, and initializes $Q(o,a) = V(o)$ for every action. Thus, the agent starts with no preference among actions. Because V is fit by Monte Carlo regression on full-episode returns, every transition receives a target computed directly from its observed return, and failed episodes still contribute by indicating which states lead to low return.

\textbf{Behavior-cloning the actor.} We use a state-based advice policy $\pi_{\text{advice}}(a \mid s)$ for this method, since operating over the full environment state makes it straightforward to construct controlled perturbations of individual sub-skills. We roll out $\pi_{\text{advice}}$ to collect demonstrations, keeping all episodes regardless of return — unlike the filtering used in Section~\ref{sec:warmstart}, since here we are not directly imitating the advice's successes, only using its rollouts to establish a state distribution to learn from. From these demonstrations, we train an actor $\pi_\theta(a \mid o)$, operating over the agent's observation $o$, to imitate the advice's actions by minimizing a cross-entropy loss with an entropy bonus to prevent premature collapse to a deterministic policy:
$
L_{\text{BC}} = L_{\text{CE}} - \alpha_H \cdot H(\pi_\theta)
$
where $H(\pi_\theta)$ is the entropy of the actor's action distribution and $\alpha_H$ is a weighting coefficient. 

Only the actor is trained at this stage; the critic is untouched. The advice acts on privileged state, whereas the agent acts only on observations. Cloning the advice into an observation-based actor ensures that the value targets fit below come from a policy that sees the same inputs as the agent.

\textbf{Rolling out the cloned actor.} We freeze the actor's parameters and roll it out under its own induced behavior, rather than the original advice policy, to collect a fresh set of trajectories. This ensures the value function we fit next is calibrated to the observation distribution the actor — and eventually the Q-network initialized from it — will actually need to generalize over, including any observations the actor visits due to imperfect imitation of the advice.

\textbf{Fitting the critic.} For every transition in the actor's rollouts, we compute the Monte Carlo return-to-go, $G_t = \sum_{k=0}^{T-t} \gamma^k r_{t+k}$, and fit a critic $V(o;\phi)$ over observations via regression against these returns:
$
L_V = \frac{1}{B}\sum \left(V(o;\phi) - G_t\right)^2
$.
Note that $V$ estimates the value of the cloned, possibly suboptimal actor and it serves only as an initialization that online TD learning subsequently corrects.

\textbf{Warmstarting the Q-network from the critic.} We copy the critic's convolutional trunk directly into the Q-network's trunk, and broadcast the critic's scalar value head across every action to initialize the Q-network's output layer:
$
W_Q \leftarrow \mathbf{1}_{|A|} \otimes w_V, \qquad b_Q \leftarrow \mathbf{1}_{|A|} \otimes b_V
$
so that $Q(o,a;\theta) = V(o)$ for every action $a$ at initialization. Standard exploration-advice DQN (Section~\ref{sec:exploration_advice}) then proceeds from this initialization, using $\pi_{\text{advice}}$ for the advice component of the exploration split. We use the original state-based advice, rather than the cloned actor, for exploration so that each perturbation affects exactly the intended sub-skills; imitation errors in the cloned actor would otherwise blur the controlled degradations.

\begin{algorithm}
{\small
    \caption{Value Warmstart for Suboptimal Advice}
    \label{alg:value-warmstart}
    \KwIn{State-based advice policy $\pi_{\text{advice}}(a \mid s)$, BC epochs $N$, critic epochs $K$, entropy weight $\alpha_H$, discount $\gamma$, batch size $B$, number of actions $|A|$}
    \KwOut{Warmstarted Q-network $Q(o,a;\theta)$}
    \BlankLine
    Initialize actor $\pi_\theta(a \mid o)$, critic $V(o;\phi)$, Q-network $Q(o,a;\theta)$\;
    \BlankLine
    \tcp{Phase 1 -- Collect demonstrations from the advice policy; keep every episode}
    Roll out $\pi_{\text{advice}}$ in the environment; store all transitions $(s_t, a_t)$ in $\mathcal{D}_{\text{demo}}$\;
    \BlankLine
    \tcp{Phase 2 -- Behavior cloning: train actor on expert actions}
    \For{$n = 1$ \KwTo $N$}{
        \ForEach{mini-batch $(o, a_E)$ of size $B$ from $\mathcal{D}_{\text{demo}}$}{
            $L_{\text{BC}} \gets L_{\text{CE}}(\pi_\theta(o), a_E) - \alpha_H \cdot H(\pi_\theta(\cdot \mid o))$\;
            Update $\theta$ by minimizing $L_{\text{BC}}$\;
        }
    }
    \BlankLine
    \tcp{Phase 3 -- Collect actor rollouts; fit the critic}
    Freeze actor parameters $\pi_\theta$\;
    Roll out $\pi_\theta$ in the environment; store transitions $(o_t, r_t)$ in $\mathcal{D}_{\text{actor}}$\;
    Compute $G_t \gets \sum_{k=0}^{T-t} \gamma^k r_{t+k}$ for every transition in $\mathcal{D}_{\text{actor}}$\;
    \For{$k = 1$ \KwTo $K$}{
        \ForEach{mini-batch $(o, G)$ of size $B$ from $\mathcal{D}_{\text{actor}}$}{
            $L_V \gets \frac{1}{B}\sum \left(V(o;\phi) - G\right)^2$\;
            Update $\phi$ by minimizing $L_V$\;
        }
    }
    \BlankLine
    \tcp{Phase 4 -- Warmstart the Q-network from the fitted critic}
    Copy $V$'s trunk into $Q$'s trunk\;
    $W_Q \gets \mathbf{1}_{|A|} \otimes w_V$; \quad $b_Q \gets \mathbf{1}_{|A|} \otimes b_V$ \tcp{broadcast value head across all actions}
    \Return{$Q(o,a;\theta)$} \tcp{$Q(o,a;\theta) = V(o)\ \forall a$ at initialization}
}
\end{algorithm}


\section{Experimental Results}
\label{sec:result}

\subsection{DQN Warmstart}

\textbf{Environment} We evaluate the first method on the MiniGrid environment suite. We consider \textit{Empty}, a navigation task requiring the agent to reach a goal square, and three object interaction tasks: \textit{Unlock}, which requires picking up a key and unlocking a door; \textit{UnlockPickup}, which additionally requires dropping the key and retrieving a box from the next room; and \textit{BlockedUnlockPickup}, which additionally requires unblocking the door before completing \textit{UnlockPickup}. All methods act from observations rather than the underlying environment state.

\begin{figure}[H]
    \centering
    \begin{subfigure}[b]{0.23\textwidth}
        \centering
        \includegraphics[width=\textwidth]{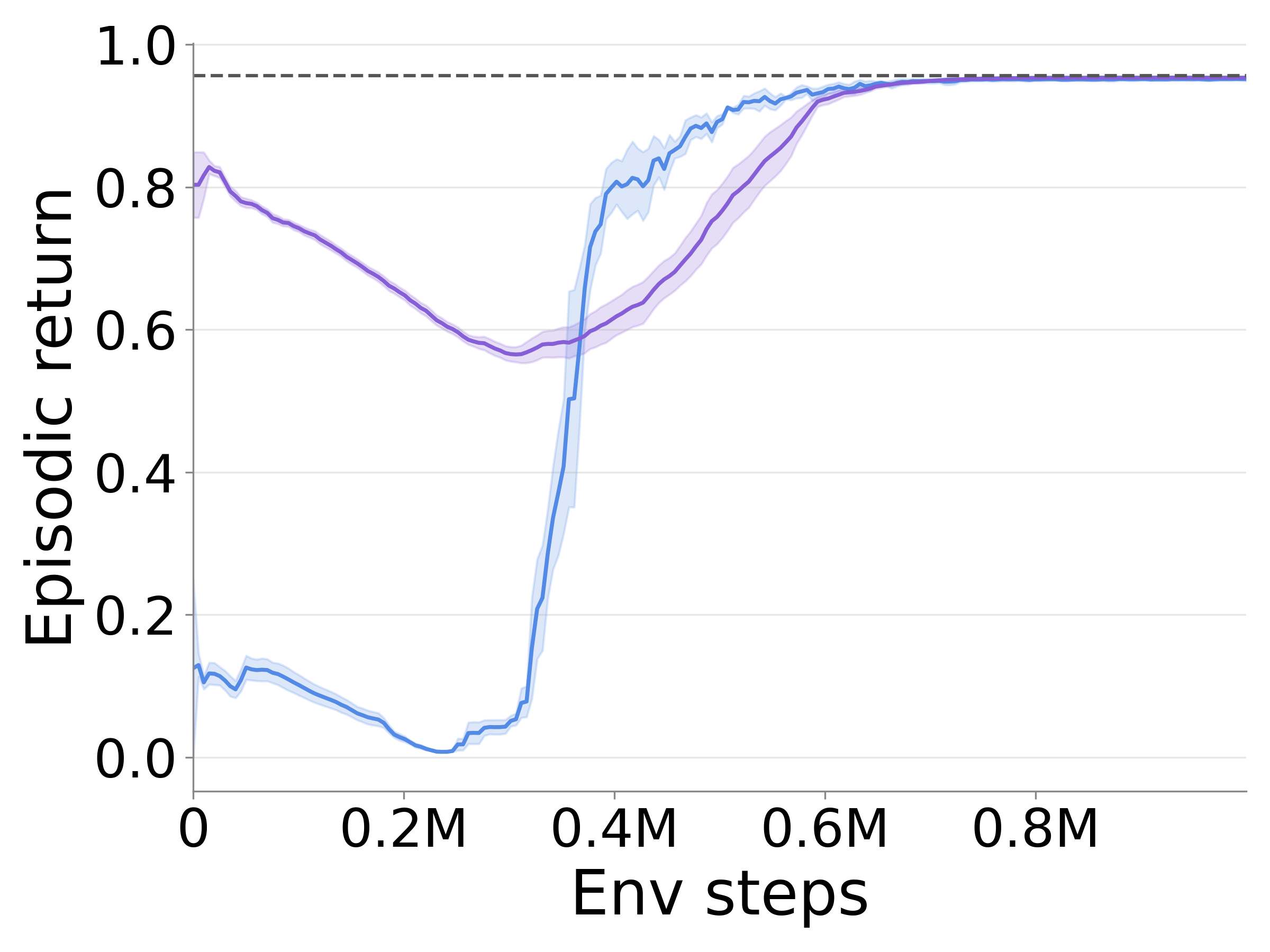}
        \caption{Empty 16x16}
        \label{fig:empty_16x16}
    \end{subfigure}
    \hfill
    \begin{subfigure}[b]{0.23\textwidth}
        \centering
        \includegraphics[width=\textwidth]{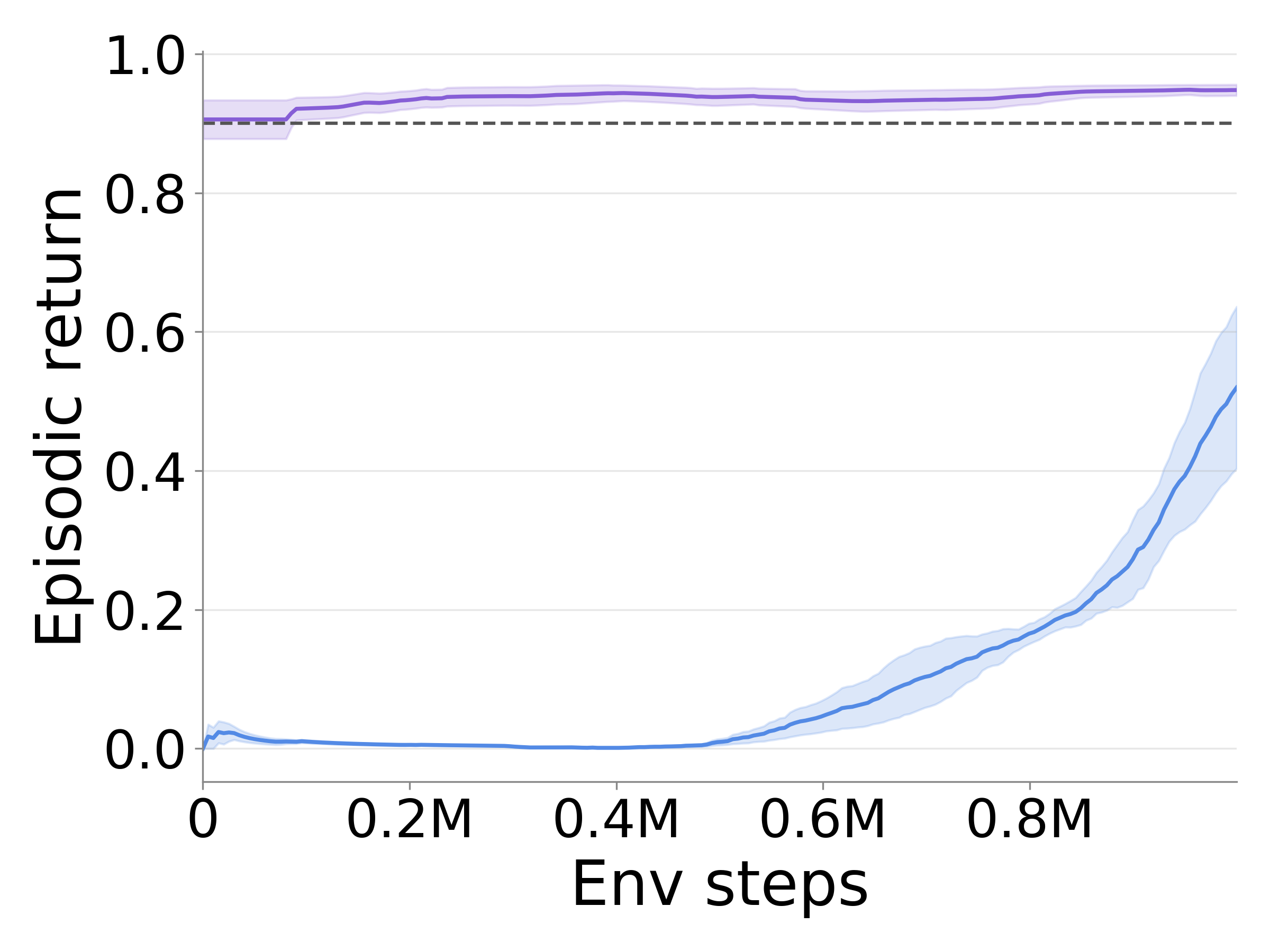}
        \caption{Unlock}
        \label{fig:unlock}
    \end{subfigure}
    \hfill
    \begin{subfigure}[b]{0.23\textwidth}
        \centering
        \includegraphics[width=\textwidth]{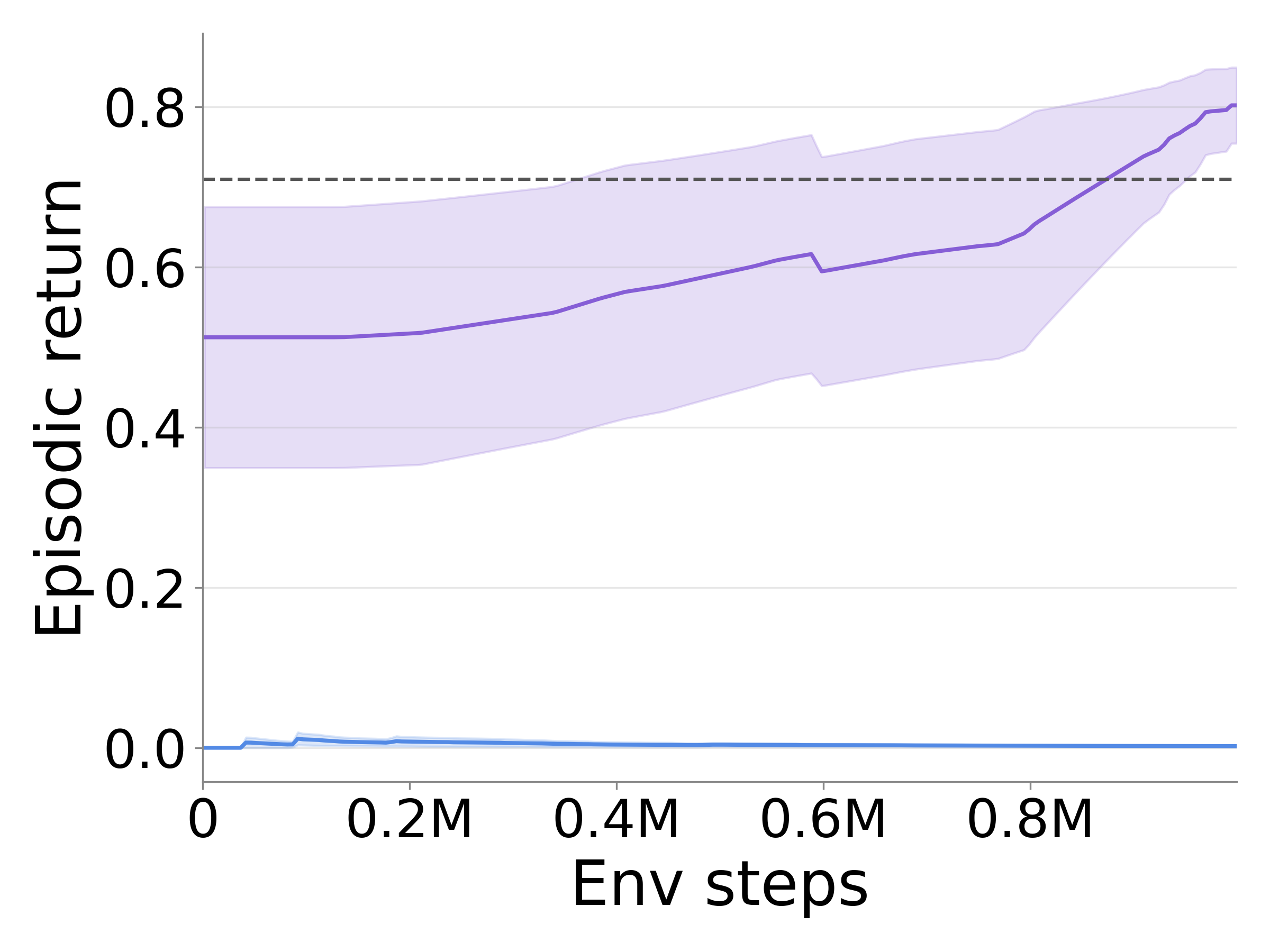}
        \caption{Unlock Pickup}
        \label{fig:unlock_pickup}
    \end{subfigure}
    \hfill
    \begin{subfigure}[b]{0.23\textwidth}
        \centering
        \includegraphics[width=\textwidth]{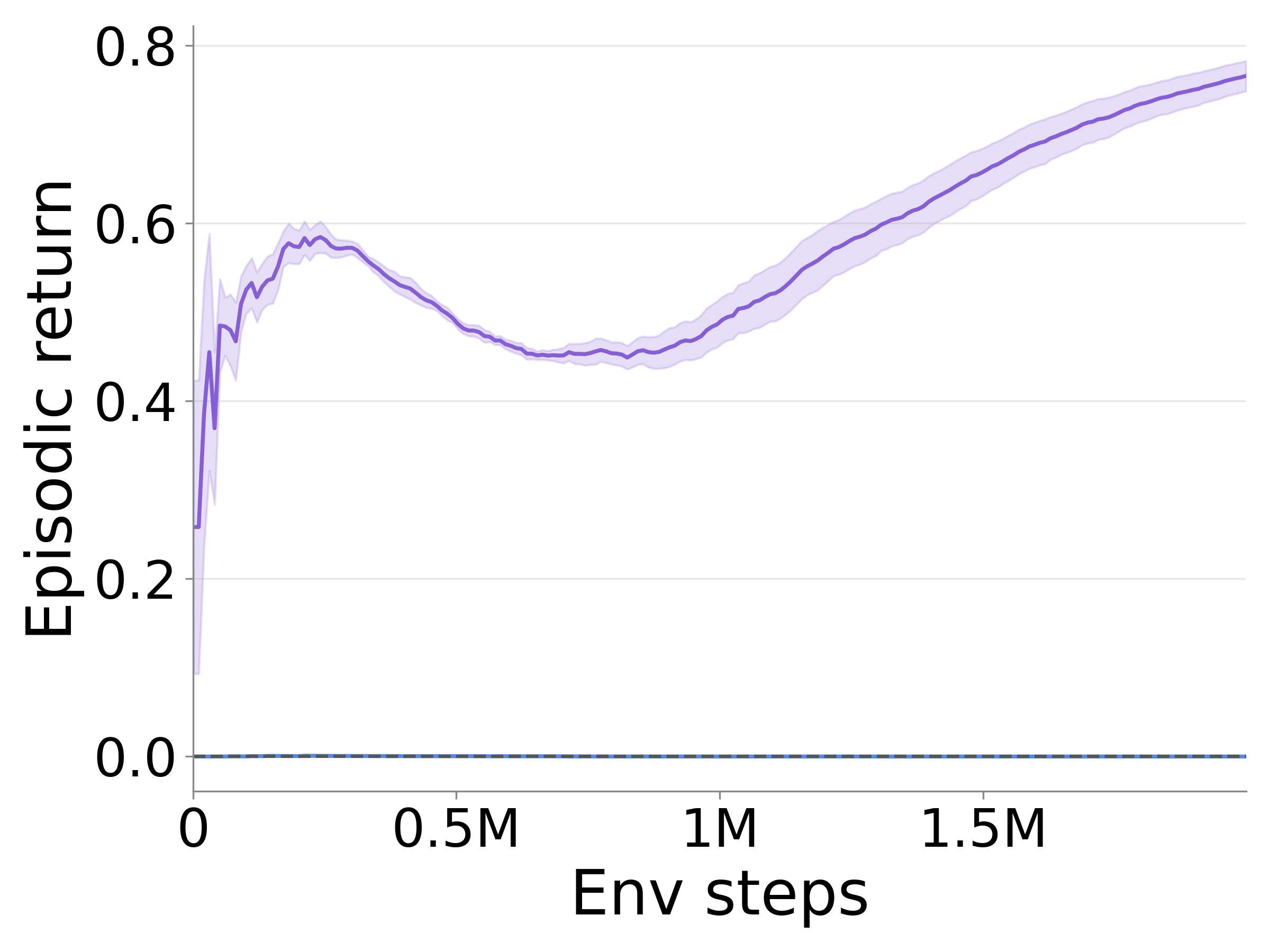}
        \caption{Blocked Unlock Pickup}
        \label{fig:blocked_unlock_pickup}
    \end{subfigure}

    \vspace{0.5em}

    \includegraphics[width=0.7\textwidth]{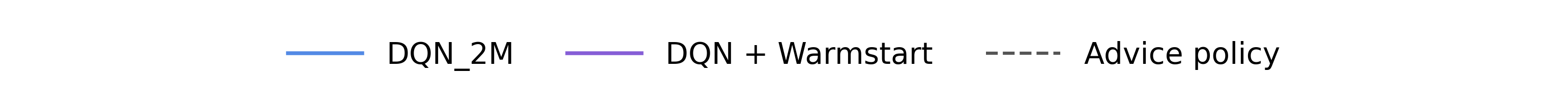}

    \caption{DQN Exploration Advice Warmstart. Curves show the mean over 5 seeds; shaded regions denote standard error.}
    \label{fig:results_optimal}
\end{figure}

\textbf{Results.} Figure ~\ref{fig:results_optimal} shows warmstarting consistently improves early learning over standard DQN, with the largest gains on Unlock, Unlock Pickup and Blocked Unlock Pickup, where DQN remains near zero for most of training. On Empty, warmstart reaches near-optimal performance substantially earlier despite an initial drop from online updates. Overall, the results suggest that advice-guided warmstarting is most beneficial when successful behavior is difficult to discover through exploration.

\subsection{Value Warmstart using Suboptimal Advice}
\label{}

We evaluate the second method on UnlockPickup, an eight-sub-skill task spanning object search, pickup, navigation, unlocking, and placement (Table~\ref{tab:subskills}). This task provides a controlled but nontrivial testbed with a subgoal structure long enough that individual sub-skill failures produce distinguishable effects on learning.

\textbf{Constructing Suboptimal Advice.} LLM-generated advice policies are sometimes suboptimal. With only egocentric, partial observations, the LLM must write rules over a limited view of the environment, and the resulting policies occasionally exhibit degenerate behavior such as moving in circles, and oscillating between actions(Appendix ~\ref{sec:obs-based_suboptimal}). However, this suboptimality is difficult to quantify as it is entangled across the policy and cannot easily be attributed to a single cause. We therefore use state-based advice to be able to quantify suboptimality.

We begin with an LLM-generated optimal state-based advice policy consisting of conditional rules for each sub-skill (Table~\ref{tab:subskills}). We perturb five of these -- locate key, pick up key, navigate to door, toggle door, and locate box -- and replace these with random actions, while keeping the rest of the sub-skills unchanged. We consider one-skill perturbations, where a single sub-skill is replaced, and two-skill perturbations, where two are replaced simultaneously (Table~\ref{tab:skill-perturbations}). This provides a labeled space of failure types, rather than an unstructured notion of advice quality, letting us measure both recovery from individual sub-skill failures and their compounding effects.

\begin{table}[h]
    \centering
    \caption{The eight sub-skills of the optimal state-based UnlockPickup policy.}
    \label{tab:subskills}
    \begin{tabular}{cl}
        \toprule
        \textbf{Sub-skill} & \textbf{Condition $\rightarrow$ Action} \\
        \midrule
        Locate Key      & $\lnot key \land \lnot open \land \lnot key\_ahead \rightarrow$ \texttt{go\_to(key)} \\
        Pick Up Key     & $\lnot key \land \lnot open \land key\_ahead \rightarrow$ \texttt{pickup} \\
        Toggle Door     & $door\_ahead \land \lnot open \rightarrow$ \texttt{toggle} \\
        Nav.\ to Door   & $key \land \lnot open \rightarrow$ \texttt{go\_to(door)} \\
        Nav.\ to Drop   & $key \land open \land \lnot at\_drop \rightarrow$ \texttt{go\_to\_drop\_spot} \\
        Drop Key        & $key \land open \land at\_drop \rightarrow$ \texttt{drop} \\
        Locate Box      & $\lnot box \rightarrow$ \texttt{go\_to(box)} \\
        Pick Up Box     & $box\_ahead \land \lnot box \rightarrow$ \texttt{pickup} \\
        \bottomrule
    \end{tabular}
\end{table}

\begin{table}[h]
    \centering
    \caption{UnlockPickup sub-skill perturbation variants. \textcolor{red}{R} indicates the sub-skill is replaced with random actions; \checkmark indicates it is left optimal.}
    \label{tab:skill-perturbations}
    \resizebox{\columnwidth}{!}{%
    \begin{tabular}{l|c|c|c|c|c}
        \toprule
        \textbf{Variant} & \textbf{A: Locate Key} & \textbf{B: Pick Up Key} & \textbf{C: Nav.\ to Door} & \textbf{D: Toggle Door} & \textbf{E: Locate Box} \\
        \midrule
        Optimal & \checkmark & \checkmark & \checkmark & \checkmark & \checkmark \\
        \midrule
        \multicolumn{6}{l}{\textit{One skill perturbed}} \\
        \midrule
        A: Random To Key & \textcolor{red}{R} & \checkmark & \checkmark & \checkmark & \checkmark \\
        B: Random Key Pickup & \checkmark & \textcolor{red}{R} & \checkmark & \checkmark & \checkmark \\
        C: Random To Door & \checkmark & \checkmark & \textcolor{red}{R} & \checkmark & \checkmark \\
        D: Random Door Toggle & \checkmark & \checkmark & \checkmark & \textcolor{red}{R} & \checkmark \\
        E: Random To Box & \checkmark & \checkmark & \checkmark & \checkmark & \textcolor{red}{R} \\
        \midrule
        \multicolumn{6}{l}{\textit{Two skills perturbed}} \\
        \midrule
        A+B & \textcolor{red}{R} & \textcolor{red}{R} & \checkmark & \checkmark & \checkmark \\
        A+C & \textcolor{red}{R} & \checkmark & \textcolor{red}{R} & \checkmark & \checkmark \\
        A+D & \textcolor{red}{R} & \checkmark & \checkmark & \textcolor{red}{R} & \checkmark \\
        A+E & \textcolor{red}{R} & \checkmark & \checkmark & \checkmark & \textcolor{red}{R} \\
        B+C & \checkmark & \textcolor{red}{R} & \textcolor{red}{R} & \checkmark & \checkmark \\
        B+D & \checkmark & \textcolor{red}{R} & \checkmark & \textcolor{red}{R} & \checkmark \\
        B+E & \checkmark & \textcolor{red}{R} & \checkmark & \checkmark & \textcolor{red}{R} \\
        C+D & \checkmark & \checkmark & \textcolor{red}{R} & \textcolor{red}{R} & \checkmark \\
        C+E & \checkmark & \checkmark & \textcolor{red}{R} & \checkmark & \textcolor{red}{R} \\
        D+E & \checkmark & \checkmark & \checkmark & \textcolor{red}{R} & \textcolor{red}{R} \\
        \bottomrule
    \end{tabular}%
    }
\end{table}

\textbf{Baselines and Evaluation}. We compare three conditions: (1) an unaided DQN agent trained without any advice, (2) the advice policy itself, executed directly with no learning, and (3) our method, which warmstarts and continues to guide a DQN agent using the (possibly perturbed) advice policy. Comparing against condition (2) isolates how much of our method's performance comes from the RL agent actually improving on the advice, rather than simply reproducing it. We report episodic return through the training run.

\begin{figure}[h]
    \centering
    \begin{subfigure}[b]{0.45\textwidth}
        \centering
        \includegraphics[width=\textwidth]{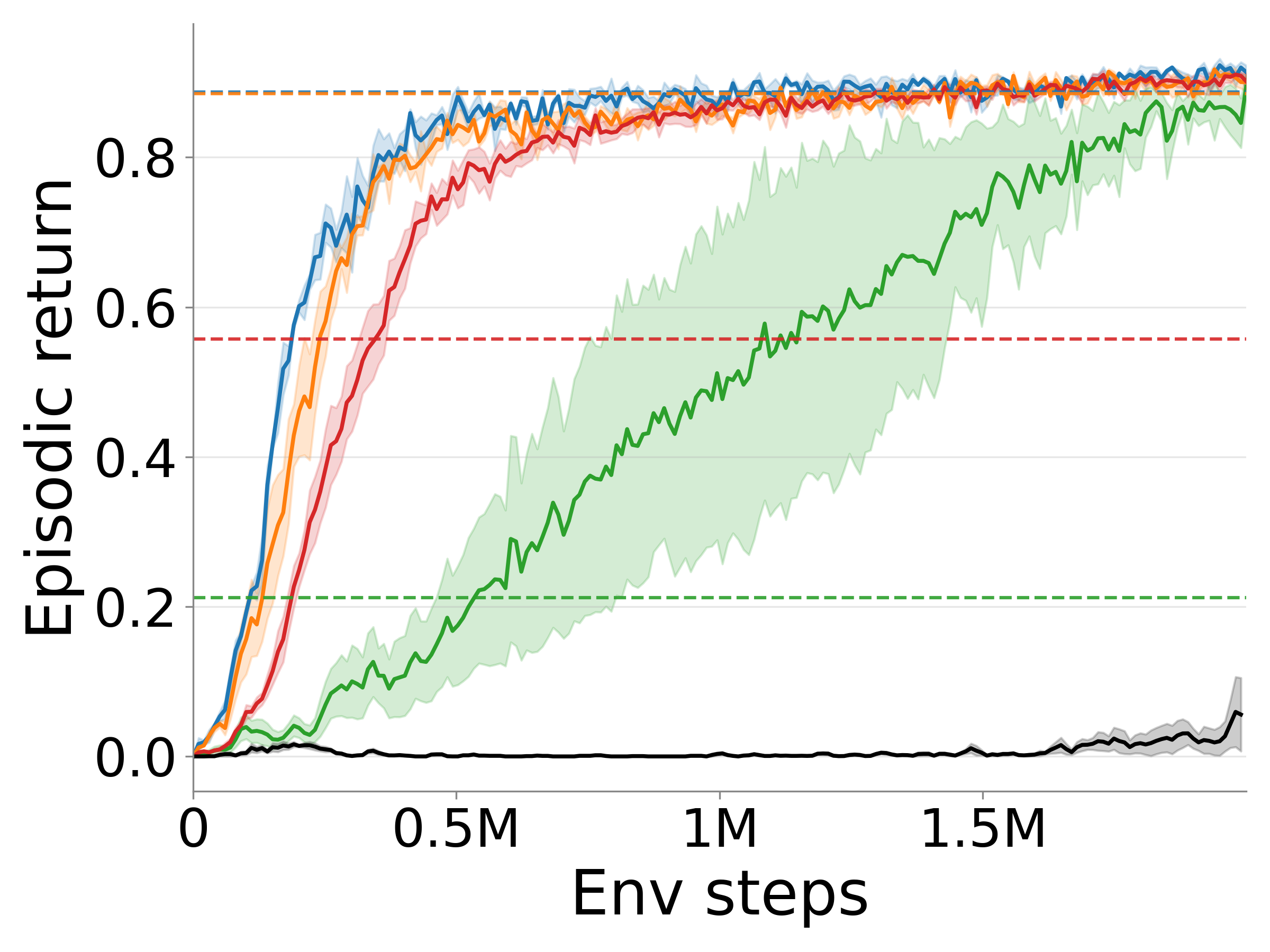}
        \label{fig:empty_6x6}
    \end{subfigure}
    \hfill
    \begin{subfigure}[b]{0.45\textwidth}
        \centering
        \includegraphics[width=\textwidth]{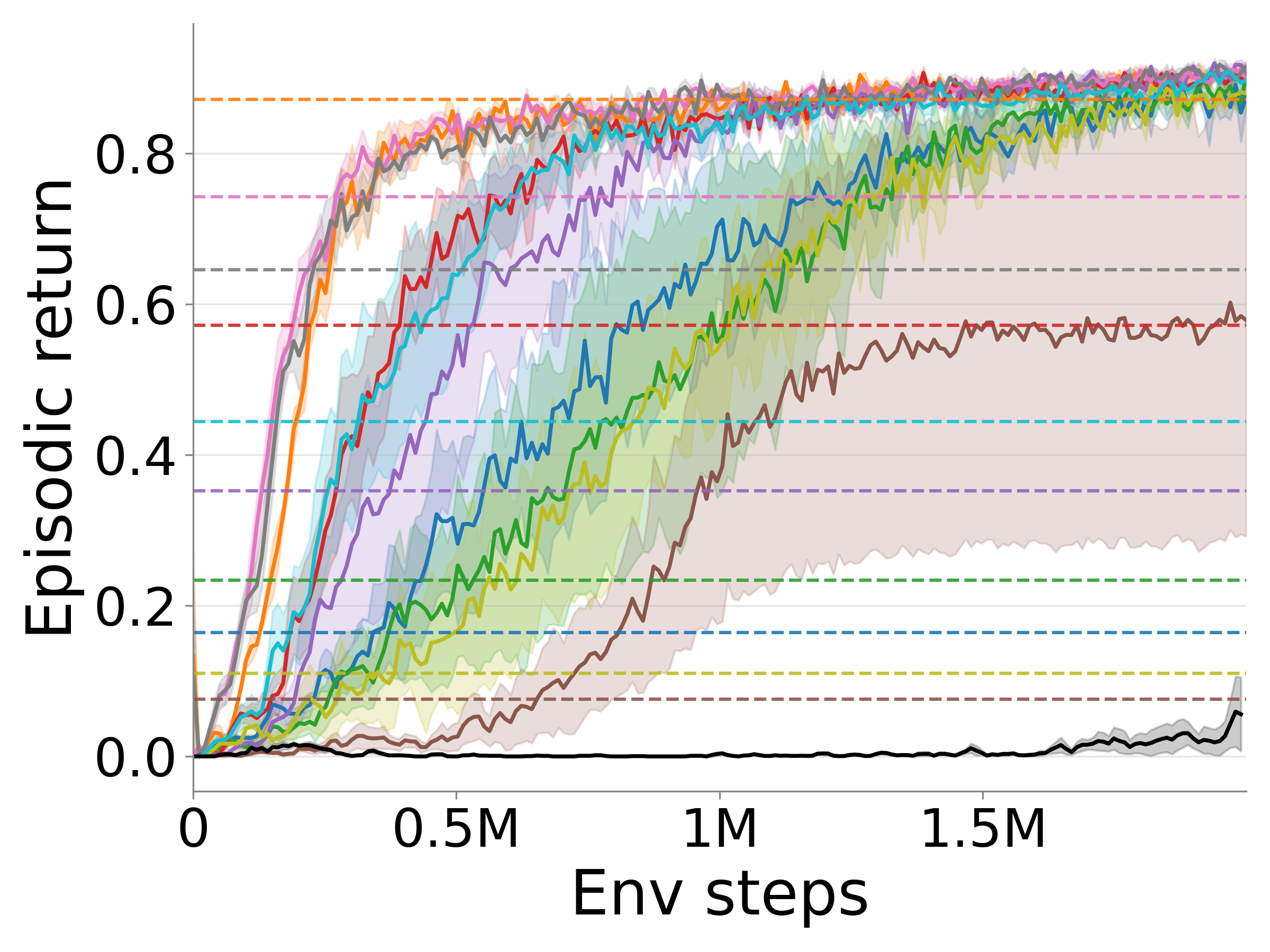}
        \label{fig:empty_16x16}
    \end{subfigure}
    \vspace{0.5em}
 
    \centering
    \begin{subfigure}[b]{0.8\textwidth}
        \centering
        \includegraphics[width=\textwidth]{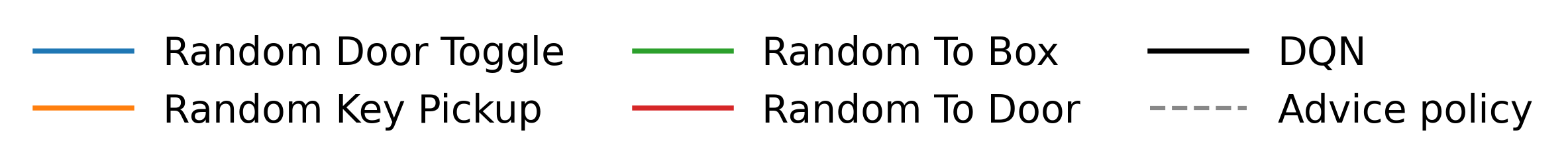}
        \caption{One Skill Perturbed}
        \label{fig:empty_6x6}
    \end{subfigure}
    \hfill
    \begin{subfigure}[b]{1.0\textwidth}
        \centering
        \includegraphics[width=\textwidth]{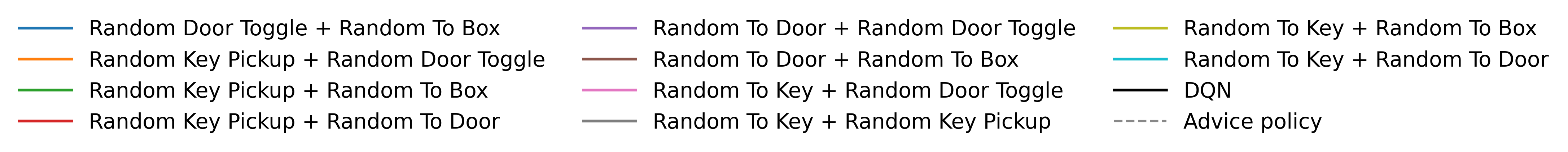}
        \caption{Two Skills Perturbed}
        \label{fig:empty_16x16}
    \end{subfigure}
    \caption{DQN Value Warmstart. Curves show the mean over 5 seeds; shaded regions denote standard error.}
    \label{fig:results_suboptimal}
\end{figure}

\textbf{Results.}\label{sec:value_warmstart_results} Figure~\ref{fig:results_suboptimal} shows episodic return through training for the one-skill and two-skill perturbed conditions. Dashed lines show the return of each perturbed advice policy executed directly without learning. Unaided DQN remains near zero throughout.

In the one-skill condition, Random Door Toggle and Random Key Pickup rise fastest, approaching a return of about 0.9 within roughly 0.5M steps. These perturbations barely degrade the advice itself (advice return $\approx 0.88$), since random actions near the key or door can still complete the sub-skill. Random To Door follows a similar trajectory with a short delay, despite its advice reaching only about 0.56. Random To Box is the clear outlier: its advice achieves roughly 0.2, and the agent improves much more slowly, reaching comparable return only near the end of training. In both Random To Door and Random To Box, the learned agent ultimately exceeds the return of its advice policy.

The two-skill condition shows a wider spread of recovery speeds. Pairings involving door toggling or key pickup recover fastest, while those involving box-finding take longer. Random To Door + Random To Box is the slowest, plateauing around 0.6 with high variance. Even so, every variant ends well above unaided DQN, and most reach high return by around 1M--1.5M steps. The results suggest that the identity of the perturbed sub-skills matters, with box-finding producing the largest degradation in learning speed.

For comparison, warmstarting directly from the same perturbed advice (Method 1; Appendix~\ref{sec:warmstart_state-based_suboptimal}) initially inherits the advice's behavior but then collapses to near-zero return between 0.5M and 1M steps, recovering only to roughly 0.4--0.7 by 2M steps in the one-skill condition. Value warmstart shows no such collapse, consistent with its action-agnostic initialization (Section~\ref{sec:warmstart_suboptimal}).


\section{Conclusion and Outlook}
\label{sec:conclusion}

We address the problem of collecting demonstrations cheaply and learning effectively from the suboptimality they sometimes contain. We use an LLM to translate an informal task description into an executable RLang advice policy, which both warmstarts a DQN agent and guides exploration. We introduce value warmstart, which learns from the advice policy's induced behavior without copying its actions. When one or two sub-skills of an LLM-generated advice policy are replaced with random behavior, value warmstart still substantially accelerates learning, while an unaided baseline fails to learn the task.

\textbf{Extending to robot manipulation.} Because RLang programs can be written over any set of skills and predicates, our methods extend naturally to robot manipulation. An LLM could write an RLang advice policy over a robot's skill library, such as those used in~\cite{ahn2022saycan}, and over predicates computed by a perception model. An RL agent selecting among the same skills could then be warmstarted and guided by this advice policy, with task completion as the sparse reward. Because the robot operates in 3-dimensional space, advice failures would arise naturally, for example when the perception model mislabels an object or a skill fails to execute, making this a realistic test for value warmstart.



\clearpage


\bibliography{main}  

\clearpage

\appendix

\section{Advice Generation Pipeline}
\label{sec:advice_generation_pipeline}

\begin{figure}[h]
    \centering
    \begin{subfigure}[b]{0.59\textwidth}
        \centering
        \includegraphics[width=\textwidth]{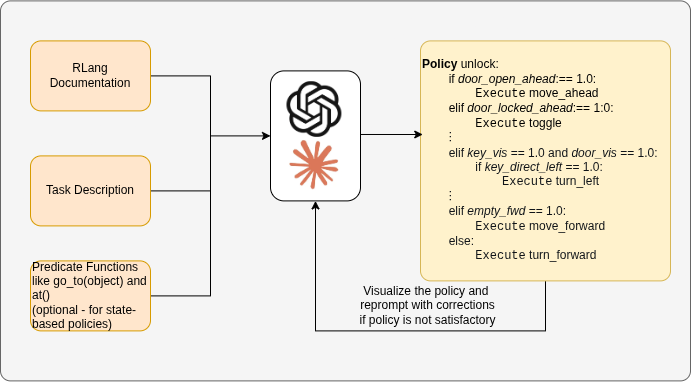}
        \caption{Advice Generation Pipeline}
        \label{fig:advice_generation_pipelines}
    \end{subfigure}
    \hfill
    \begin{subfigure}[b]{0.31\textwidth}
        \centering
        \includegraphics[width=\textwidth]{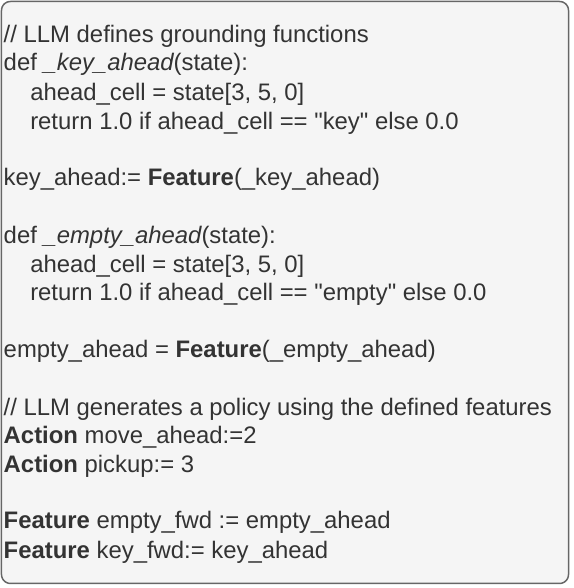}
        \caption{RLang Features Definitions}
        \label{fig:rlang}
    \end{subfigure}

    \caption{(a) An LLM is prompted with RLang documentation, a task description, and, for state-based advice, predicate functions, and returns an executable RLang advice policy $\pi_{advice}$. (b) Excerpt of LLM-generated RLang Features}
    \label{fig:advice_generation}
\end{figure}

We use Claude Sonnet 5.5 to generate all advice policies. The prompt contains RLang's documentation (its grammar and annotated example programs), a description of the task (its objective, observation and action spaces, and optionally a textual strategy), and, for state-based advice, predicate functions such as \texttt{at()} and \texttt{go\_to()} (Figure~\ref{fig:advice_generation_pipelines}). We roll out each generated policy and inspect its behavior; if it fails the task or behaves degenerately, we re-prompt the LLM with natural-language corrections describing the observed failure. The number of re-prompts needed varies by task: Empty required none, while BlockedUnlockPickup, the longest-horizon task, required around six. Figure~\ref{fig:rlang} and Listing~\ref{lst:rlang_features} show representative LLM-generated features, and Listing~\ref{lst:rlang_policy} shows the resulting policy for Unlock.

\label{lst:features}
\begin{lstlisting}[
    language=Python,
    caption={Three representative LLM-generated feature functions for the Unlock environment, showing how each is computed from the $7\times7$ egocentric observation before being bound to an RLang \texttt{Feature}.},
    label={lst:rlang_features},
    basicstyle=\ttfamily\small,
    keywordstyle=\color{blue},
    commentstyle=\color{gray},
    stringstyle=\color{orange},
    frame=single,
    breaklines=true
]
# (1) Single-cell check: is the cell directly ahead a key?
def _key_ahead(state):
    ahead_cell = state[3, 5, 0]
    return 1.0 if ahead_cell == OBJECT_TO_IDX["key"] else 0.0
 
# (2) Single-cell check with auxiliary state: is the door
#     directly ahead both present and locked?
def _door_locked_ahead(state):
    ahead_cell = state[3, 5, 0]
    ahead_state = state[3, 5, 2]
    return 1.0 if (ahead_cell == OBJECT_TO_IDX["door"]
                   and ahead_state == DOOR_LOCKED) else 0.0
 
# (3) Full-window scan: is a key visible anywhere in view,
#     regardless of position?
def _key_visible(state):
    for i in range(7):
        for j in range(7):
            if (i, j) == (3, 6):  # skip the agent's own cell
                continue
            if state[i, j, 0] == OBJECT_TO_IDX["key"]:
                return 1.0
    return 0.0
 
# Each function is bound to an RLang Feature of the same name
Feature key_fwd         := key_ahead
Feature door_locked_fwd := door_locked_ahead
Feature key_vis         := key_visible
\end{lstlisting}

\label{lst:policy}
\begin{lstlisting}[
    caption={Condensed RLang policy for the Unlock environment, as generated by the LLM. The policy encodes a priority-ordered strategy: open the door if directly ahead, pick up the key if visible, navigate to the door once the key is held, and explore otherwise.},
    label={lst:rlang_policy},
    basicstyle=\ttfamily\small,
    keywordstyle=\color{blue},
    commentstyle=\color{gray},
    frame=single,
    breaklines=true
]
Policy unlock:
    # Step 1: If door is open ahead, move through it
    if door_open_ahead == 1.0:
        Execute move_forward
    # Step 2: Toggle door if locked or closed ahead
    elif door_locked_ahead == 1.0 or door_closed_ahead == 1.0:
        Execute toggle
    # Step 3: Pick up key if directly in front
    elif key_fwd == 1.0:
        Execute pickup
    # Step 4: Navigate to key if visible
    elif key_vis == 1.0:
        if key_direct_left == 1.0:
            Execute turn_left
        elif key_direct_right == 1.0:
            Execute turn_right
        elif empty_fwd == 1.0:
            Execute move_forward
        else:
            Execute turn_right
    # Step 5: Navigate to door if visible (key already held)
    elif door_vis == 1.0:
        if door_direct_left == 1.0:
            Execute turn_left
        elif door_direct_right == 1.0:
            Execute turn_right
        elif empty_fwd == 1.0:
            Execute move_forward
        else:
            Execute turn_right
    # Step 6: Explore - wall following fallback
    elif wall_fwd == 1.0:
        Execute turn_right
    elif empty_fwd == 1.0:
        Execute move_forward
    else:
        Execute turn_left
\end{lstlisting}

\clearpage

\section{Observation-based Suboptimal Advice}
\label{sec:obs-based_suboptimal}

\begin{figure}[h]
    \includegraphics[width=0.9\textwidth]{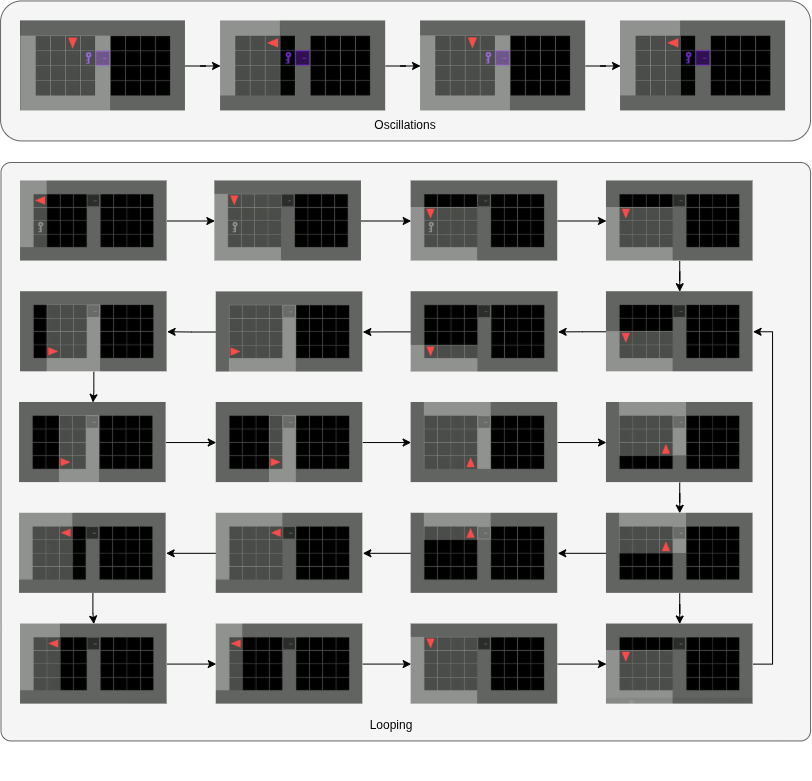}
    \caption{Suboptimal Advice Generation. Degenerate behavior of LLM-generated observation-based advice in UnlockPickup. \textbf{Top (Oscillations):} the agent alternates between the same few actions without making progress. \textbf{Bottom (Looping):} the agent follows a closed path through the room and returns to an earlier state, repeating the cycle.}
        \label{fig:suboptimal_advice_generation_examples}
\end{figure}

Figure~\ref{fig:suboptimal_advice_generation_examples} shows two failure modes we observed in LLM-generated observation-based advice: oscillating between actions and looping through the same states. Because such failures arise from many interacting rules, their cause is hard to isolate.

\begin{figure}[h]
    \centering
    \begin{subfigure}[b]{0.32\textwidth}
        \centering
        \includegraphics[width=\textwidth]{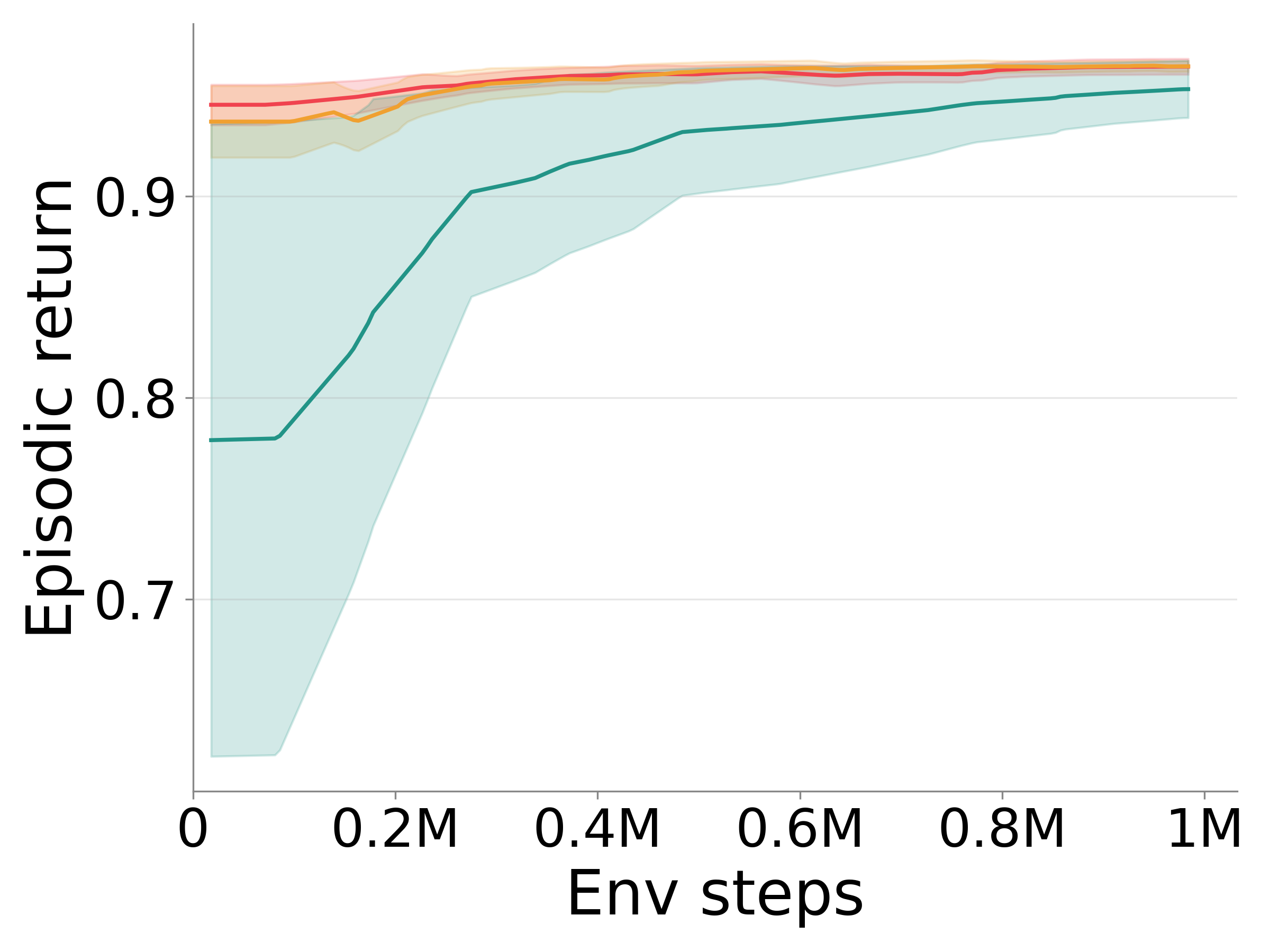}
        \caption{Empty}
        \label{fig:empty}
    \end{subfigure}
    \hfill
    \begin{subfigure}[b]{0.32\textwidth}
        \centering
        \includegraphics[width=\textwidth]{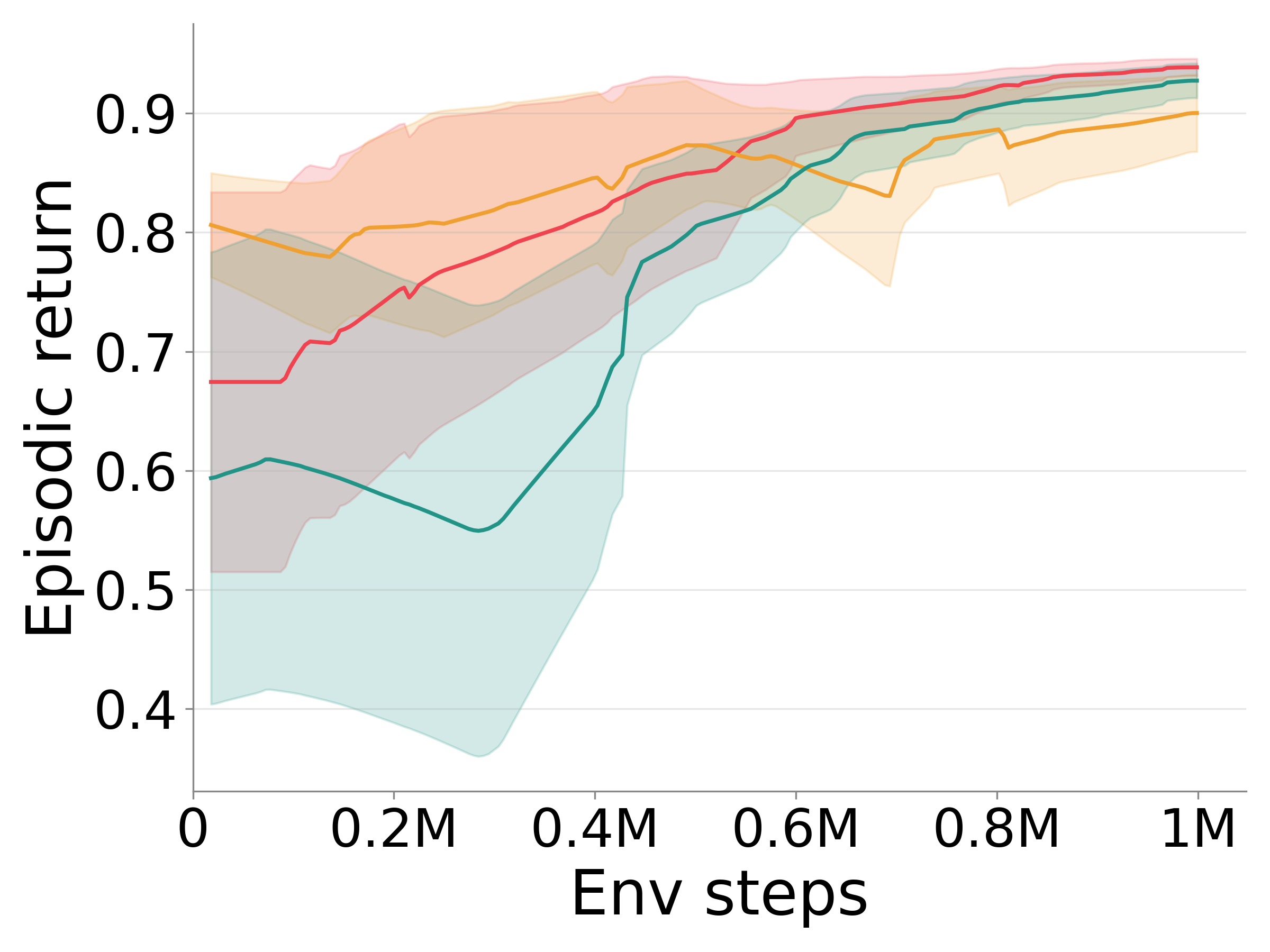}
        \caption{Unlock}
        \label{fig:Unlock}
    \end{subfigure}
    \hfill
    \begin{subfigure}[b]{0.32\textwidth}
        \centering
        \includegraphics[width=\textwidth]{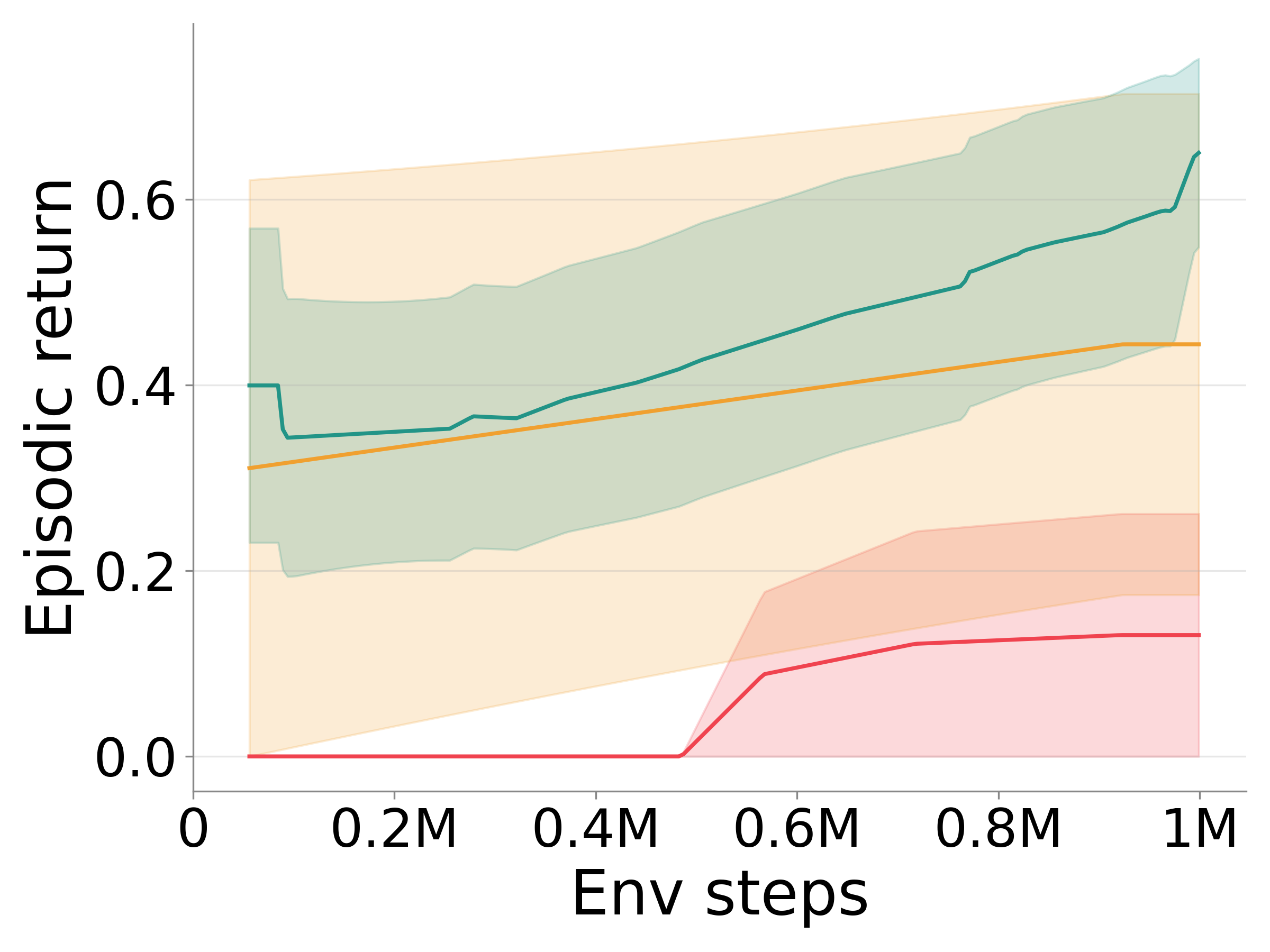}
        \caption{Unlock Pickup}
        \label{fig:unlock_pickup}
    \end{subfigure}
 
    \vspace{0.5em}

    \includegraphics[width=0.7\textwidth]{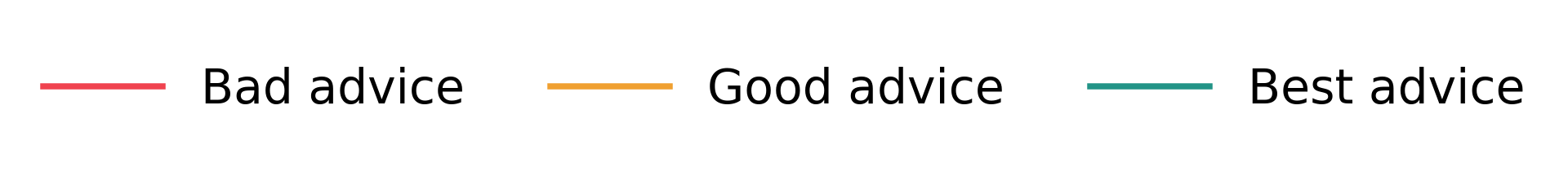}
 
    \caption{Suboptimal Observation-based Advice. Curves show the mean over 5 seeds; shaded regions denote standard error.}
    \label{fig:obs_suboptimal}
\end{figure}

Before constructing the controlled state-based degradations, we evaluated warmstarting with observation-based advice of varying quality. For each of Empty, Unlock, and UnlockPickup, we wrote three RLang advice policies by hand over egocentric observations, to mimic the range of quality we observed in LLM-generated advice:
\begin{description}
    \item[Bad] reacts only to the cell directly ahead, with no carrying-state checks, so it cannot sequence subgoals (e.g., it may toggle the door before fetching the key) and falls back on wall-following for navigation. On Empty, it ignores the goal and mostly turns in place, though it still generally reaches the goal since the start and goal positions are fixed.
    \item[Good] sequences subgoals correctly using visibility and carrying-state checks, but has no directional steering: it moves forward only when the target is visible and turns otherwise.
    \item[Best] uses the same subgoal logic as Good, plus directional features that steer the agent toward each target. On UnlockPickup, it also turns aside before dropping the key so that the key does not block the box.
\end{description}

Figure~\ref{fig:obs_suboptimal} shows warmstarting (Method 1) with each variant. On Empty and Unlock, all three variants converge to similar return within 1M steps, so advice quality mainly affects early learning. On UnlockPickup, where subgoal sequencing matters most, the ordering is preserved: the Bad variant yields almost no return until about 0.5M steps and plateaus near 0.1, while the Good and Best variants start around 0.3--0.4 and keep improving throughout training. Because these variants differ in many ways at once, their suboptimality cannot be attributed to a single cause, which motivates the controlled state-based degradations in the main text.

\clearpage

\section{Warmstart on State-Based Suboptimal Advice}
\label{sec:warmstart_state-based_suboptimal}

\begin{figure}[h]
    \centering
    \begin{subfigure}[b]{0.45\textwidth}
        \centering
        \includegraphics[width=\textwidth]{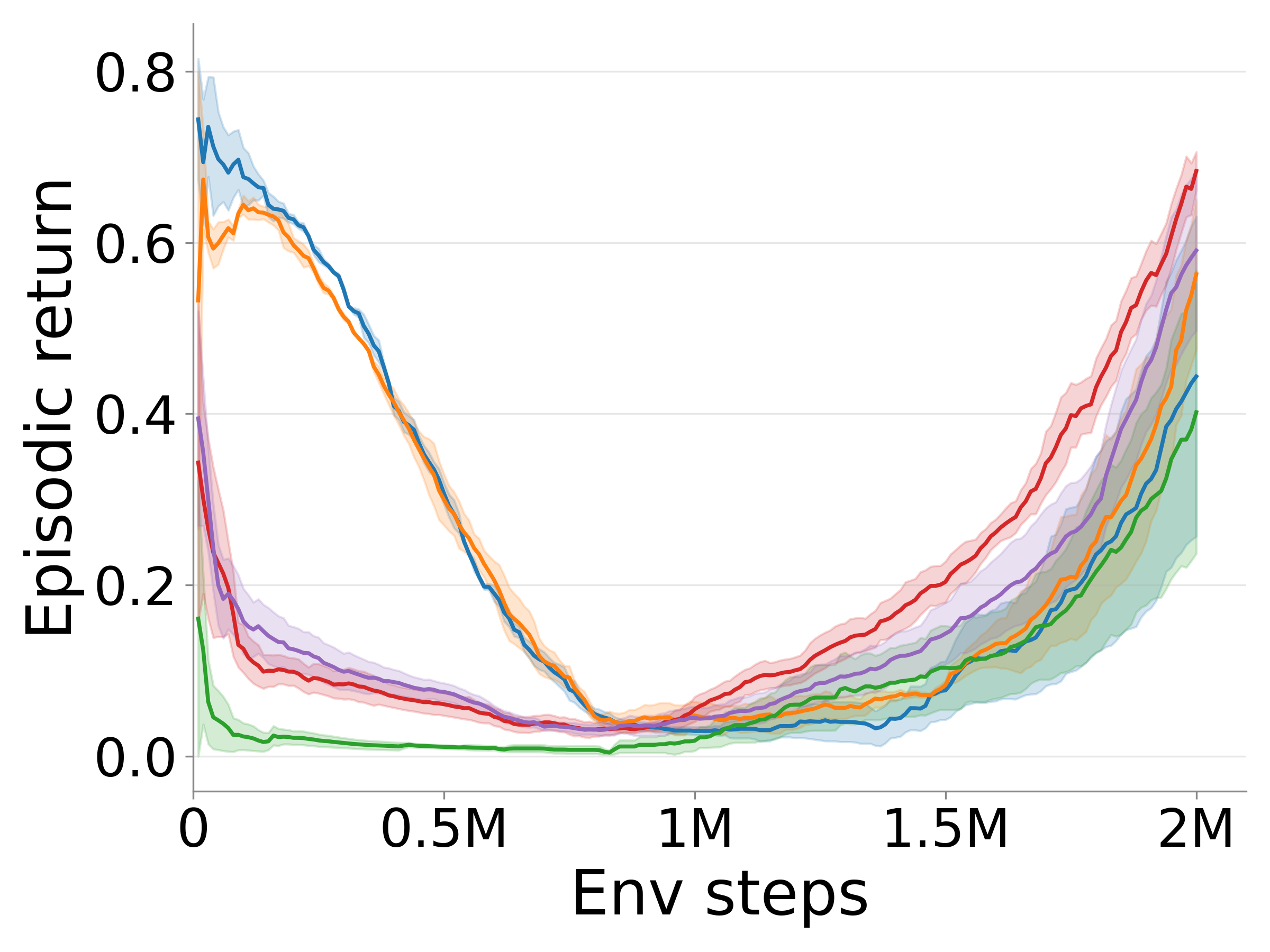}
        \label{fig:empty_6x6}
    \end{subfigure}
    \hfill
    \begin{subfigure}[b]{0.45\textwidth}
        \centering
        \includegraphics[width=\textwidth]{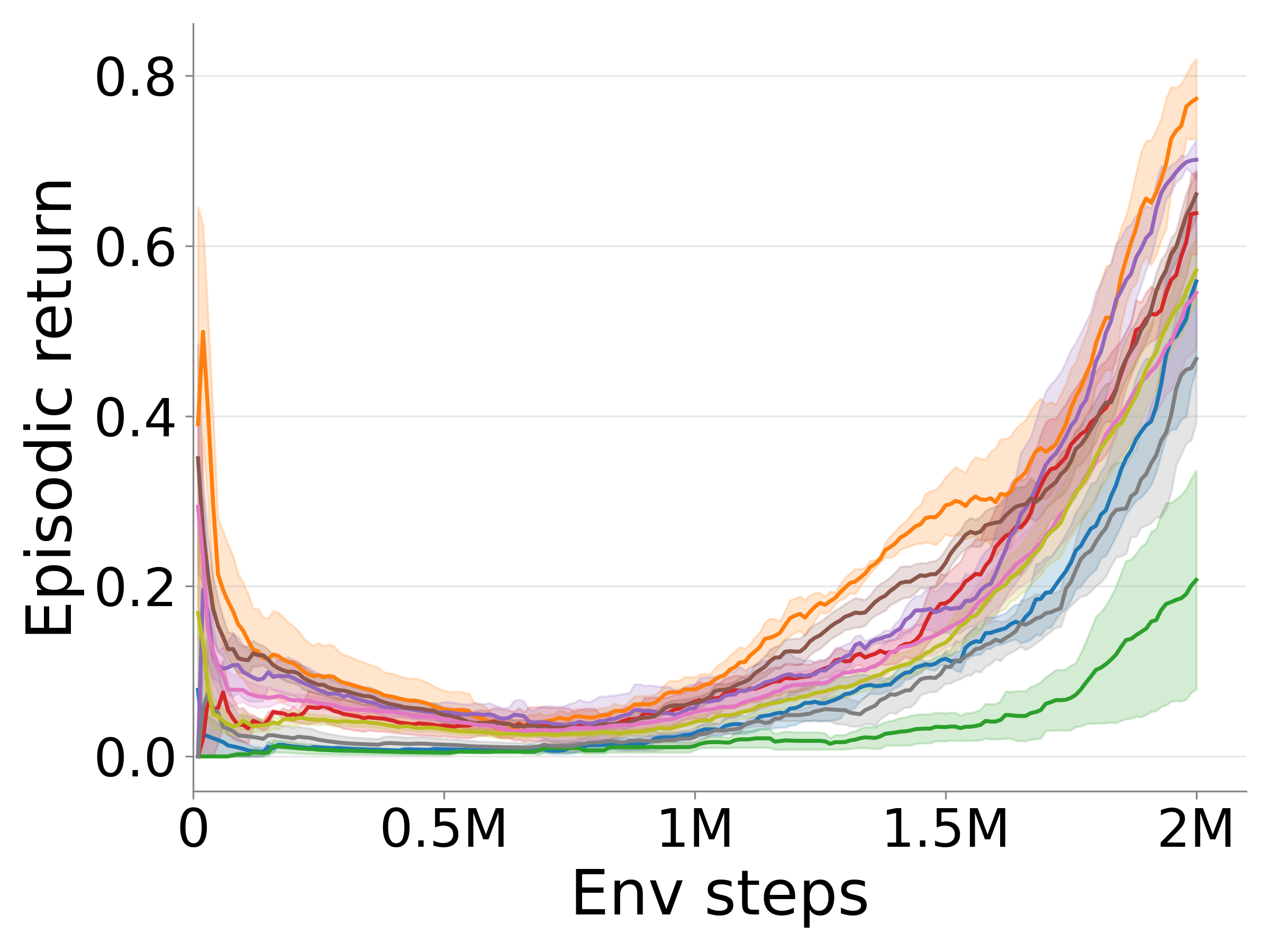}
        \label{fig:empty_16x16}
    \end{subfigure}
    \vspace{0.5em}
 
    \centering
    \begin{subfigure}[b]{0.8\textwidth}
        \centering
        \includegraphics[width=\textwidth]{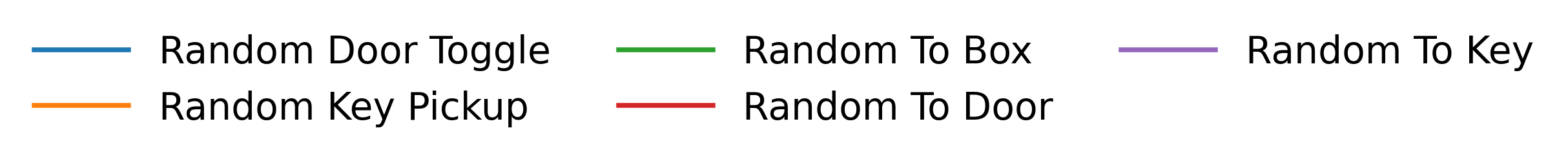}
        \caption{One Skill Perturbed}
        \label{fig:empty_6x6}
    \end{subfigure}
    \hfill
    \begin{subfigure}[b]{0.8\textwidth}
        \centering
        \includegraphics[width=\textwidth]{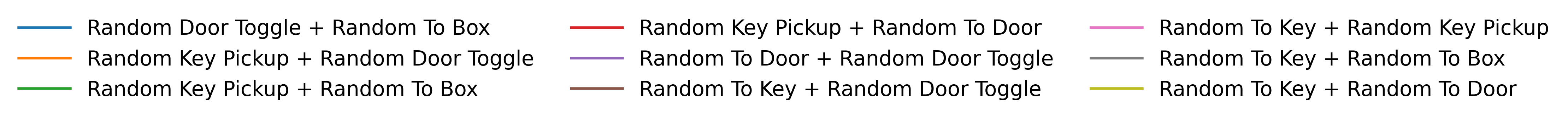}
        \caption{Two Skills Perturbed}
        \label{fig:empty_16x16}
    \end{subfigure}
    \caption{DQN Warmstart Suboptimal. Curves show the mean over 5 seeds; shaded regions denote standard error.}
    \label{fig:state-based_suboptimal_warmstart}
\end{figure}

Figure~\ref{fig:state-based_suboptimal_warmstart} shows Method 1 (Section~\ref{sec:warmstart}) trained with the same one- and two-skill perturbations as in Section~\ref{sec:value_warmstart_results}. Every variant follows the same qualitative shape. Returns start high, because the pretrained Q-network already reproduces the advice's actions and half of the agent's actions are greedy at the start of training. They then decline into a trough near zero between 0.5M and 1M steps, as online updates overwrite the imitated behavior before the agent has learned to compensate for the degraded sub-skill. Finally, they recover steadily from roughly 1M steps onward through the end of training.

Variants whose perturbation barely degrades the advice (Random Door Toggle and Random Key Pickup) start highest and take longest to decline. Random To Box starts near zero and recovers most slowly, consistent with Section~\ref{sec:value_warmstart_results}, where box-finding was also the hardest missing sub-skill to recover. The two-skill condition shows the same shape and a similar final range, except for Random Key Pickup + Random To Box, which reaches only about 0.2 by 2M steps. By contrast, value warmstart (Figure~\ref{fig:results_suboptimal}) starts near zero, since its Q-network has no action preference at initialization, and rises without collapsing, reaching about 0.9 within 0.5M steps on most one-skill variants.

\clearpage

\section{Hyperparameters}
\label{subsec:appendix_dqn_hyperparams}

Table~\ref{tab:dqn-hyperparams} lists the hyperparameters for the DQN and the proposed methods.

\begin{table}[H]
    \centering
    \caption{Hyperparameters. DQN settings are shared by all methods.}
    \label{tab:dqn-hyperparams}
    \begin{tabular}{lc}
        \toprule
        \textbf{Hyperparameter} & \textbf{Value} \\
        \midrule
        \multicolumn{2}{l}{\textit{DQN (all methods)}} \\
        Learning rate                            & $1 \times 10^{-4}$, linearly annealed \\
        Replay buffer size                       & 100{,}000 \\
        Discount factor $\gamma$                 & 0.99 \\
        Target network update rate $\tau$        & 1.0 \\
        Target network update frequency          & 1{,}000 steps \\
        Batch size                               & 32 \\
        Initial exploration $\varepsilon_{\text{start}}$ & 0.5 \\
        Final exploration $\varepsilon_{\text{end}}$     & 0.05 \\
        Exploration fraction $f$                 & 0.2 \\
        Learning starts                          & 20{,}000 steps \\
        Train frequency                          & every 4 steps \\
        Network                                  & 3 conv layers (32, 64, 64) + 512-unit FC \\
        \midrule
        \multicolumn{2}{l}{\textit{Method 1: Warmstarting}} \\
        n-step horizon $n$                       & 10 \\
        n-step loss weight $\lambda$             & 1.0 \\
        \midrule
        \multicolumn{2}{l}{\textit{Method 2: Value Warmstart}} \\
        Demonstration episodes                   & 200 \\
        BC learning rate                         & $3 \times 10^{-4}$ \\
        BC batch size                            & 64 \\
        BC epochs $N$                            & 200 \\
        Entropy weight $\alpha_H$                & 0.01 \\
        Actor rollout episodes                   & 200 \\
        Critic learning rate                     & $3 \times 10^{-4}$ \\
        Critic batch size                        & 32 \\
        Critic epochs $K$                        & 50 \\
        \bottomrule
    \end{tabular}
\end{table}

\end{document}